\documentclass[letterpaper, 10 pt, conference]{ieeeconf}
\IEEEoverridecommandlockouts                               
\usepackage{breqn}
\usepackage{cuted}
\usepackage{xcolor}
\usepackage{capt-of}
\usepackage{amsfonts}
\usepackage{hyperref}
\usepackage{multirow}
\usepackage{mathtools}
\usepackage{dblfloatfix}
\usepackage[font=small, labelfont=bf]{caption}
\usepackage[ruled, lined, linesnumbered, commentsnumbered, longend, noend]{algorithm2e}

\usepackage{fancyhdr} 
\definecolor{pink}{RGB}{255, 192, 203}

\title{\LARGE \bf Bio-Inspired Hybrid Map: Spatial Implicit Local Frames and Topological Map for Mobile Cobot Navigation}

\author{\large Tuan Dang \hspace{45pt} Manfred Huber {\footnotesize \thanks{All authors are with the Learning and Adaptive Robotics Laboratory, Department of Computer Science and Engineering, University of Texas at Arlington, Arlington, TX 76013, USA. (emails: \href{mailto:tuan.dang@uta.edu}{\text{tuan.dang@uta.edu}},  \href{mailto:huber@cse.uta.edu}{\text{huber@cse.uta.edu}})}}}

\begin{document}

\maketitle 

\thispagestyle{empty}
\pagestyle{empty}

\begin{abstract}

Navigation is a fundamental capacity for mobile robots, enabling them to operate autonomously in complex and dynamic environments. Conventional approaches use probabilistic models to localize robots and build maps simultaneously using sensor observations. Recent approaches employ human-inspired learning, such as imitation and reinforcement learning, to navigate robots more effectively. However, these methods suffer from high computational costs, global map inconsistency, and poor generalization to unseen environments. This paper presents a novel method inspired by how humans perceive and navigate themselves effectively in novel environments. Specifically, we first build local frames that mimic how humans represent essential spatial information in the short term. Points in local frames are hybrid representations, including spatial information and learned features, so-called spatial-implicit local frames. Then, we integrate spatial-implicit local frames into the global topological map represented as a factor graph. Lastly, we developed a novel navigation algorithm based on Rapid-Exploring Random Tree Star (RRT*) that leverages spatial-implicit local frames and the topological map to navigate effectively in environments. To validate our approach, we conduct extensive experiments in real-world datasets and in-lab environments. We open our source code at \href{https://github.com/tuantdang/simn}{https://github.com/tuantdang/simn}.

\end{abstract}

\section{Introduction}

Mobile collaborative robots combine the flexibility of mobile robots with the adaptability of collaborative robots, enabling them to work alongside humans in dynamic environments. This feature enhances workplace safety, making these robots essential in modern automation, particularly indoor applications. SLAM-based methods ( simultaneous localization and mapping) \cite{mur2017orb, newcombe2011dtam, schops2019bad, whelan2015elasticfusion, bloesch2018codeslam, chung2023orbeez, czarnowski2020deepfactors, dang2024v3d} are vital for allowing robots to operate autonomously. However, these methods require significant computational resources, especially in large-scale environments. Additionally, they often struggle with drift, affecting the global map's consistency across different observations.

Unlike classical SLAM methods, which rely on handcrafted feature extraction and probabilistic models, learning-based SLAM approaches utilize data-driven techniques for feature extraction, pose estimation, and loop closure detection \cite{mccormac2017semanticfusion, sucar2020nodeslam, zhi2019scenecode, sucar2021imap, johari2023eslam}. However, these learning-based methods require labeled data to train the models and often work with specific datasets, which limits their ability to adapt to new environments. Additionally, SLAM systems are typically updated at the rate of the sensors, which can lead to the accumulation of pose estimation errors over time. Map building heavily depends on sensor quality and precise calibration, making it challenging to create consistent map representations while accurately localizing robots in noisy environments. As a result, navigating in dynamic, resource-constrained settings may lead to poor planning outcomes.

\begin{figure}[t]
    \centering
    \includegraphics[width=0.99\linewidth]{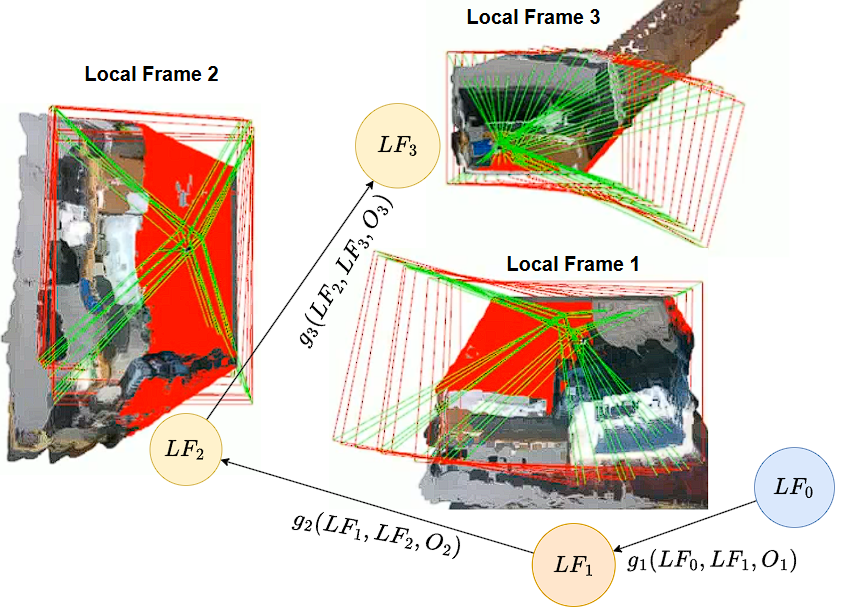}
    \caption{Bio-inspired hybrid map concept: the robot aims to continuously learn and construct precise 3D semantic local frames using spatial information and learned features for effective collision avoidance while creating a topological map connecting local frames for global guidance, facilitating navigation tasks.}
    \label{fig:concept}
    \vspace{-15 pt}
\end{figure}

When comparing sensory systems between humans and robots (e.g., LiDAR or RGB-D cameras), robots generally excel at estimating distances to obstacles. In contrast, humans often struggle to judge these distances accurately. This raises the question: why are we adept at quickly navigating complex and dynamic environments, even in previously unfamiliar settings with imprecise sensors? One possible explanation is that, unlike SLAM systems, we do not attempt to create precise maps from observations. Instead, we localize ourselves based on a few recent observations, with minimal reliance on global information. Our ability to navigate relies heavily on our understanding of relative positions between known landmarks. Specifically, the hippocampus and entorhinal cortex are involved in encoding, storing, and retrieving spatial information \cite{frank2000trajectory, buzsaki2013memory, lisman2017viewpoints}, where cognitive maps are developed to represent spatial relationships. Humans utilize landmarks as reference points to help reorient themselves and navigate in unseen environments.

We propose a hybrid mapping system inspired by how humans represent, store, and retrieve spatial information to build cognitive maps and perceive landmarks for localization. This system incorporates local frames encoded as spatial-implicit coordinates and a global map represented as a flexibly topological map. The local frames hold spatial information and learned features that can be created, stored, and retrieved during robot operations. We represent each local frame as a node in the topological map, where an arc represents the relative spatial relationship between local frames. Figure \ref{fig:concept} shows the basic concept of this hybrid representation. The approach also implements an algorithm that allows robots to automatically form new local frames by observing variances in the accumulated travel distance and viewpoints. We make contributions to this work as follows:
\begin{itemize}
    \item Constructing the spatial-implicit local frames and topological map that mimic how humans encode, store, and retrieve spatial information
    \item Building the continuous self-supervised learning models that help robots quickly adapt to new environments
    \item Developing the cognition-inspired navigation algorithm to help robot planning more efficiently
    \item Benchmarking our methods in real-world datasets and in-lab environments
\end{itemize}

\section{Related Work}

\subsection{Learned Spatial Representation}
Robots must adapt quickly to new environments, as continuous learning and lifelong learning are essential in robotic applications. Recent advancements in sim-to-real reinforcement learning and sim-to-real imitation learning approaches have received much attention from the robotic research community. These approaches allow robots to learn in simulated environments safely before transferring their learned policies to real-world applications. However, challenges remain, including the reality gap between simulated and real environments, the high computation cost of training, and re-training when changing to different environments. Instead of training robots in a specific simulated environment, we propose allowing the robot models to learn their spatial representation \cite{ zhi2019scenecode, sucar2021imap, johari2023eslam} from unseen real-time observations through continual learning in a self-supervised manner. This approach enables robots to learn spatial information more rapidly, facilitating quicker adaptation to the new environments.

\subsection{Metric vs Topological Maps}
Most work in topological mapping defines nodes as significant places or landmarks, while arcs connecting these nodes represent spatial connectivity or action sequences. Theoretically, topological maps should scale better than metric maps due to their coarse-grained nature and compact representation of large-scale scenes. However, designing a topological map can be challenging since compact representation may make it difficult to distinguish among distinctive landmarks. On the other hand, recent works on metric maps  \cite{chung2023orbeez, czarnowski2020deepfactors, dang2024v3d}, particularly in 3D-SLAM, have achieved remarkable results in creating precise large-scale maps. Nevertheless, memory and time complexity raise serious problems when the system constantly updates maps with every observation as these maps get larger. Our approach combines metric maps as local maps with topological maps, leveraging both strengths \cite{thrun2001robust}. Local frames are learned from real-time observations, represented in hybrid spatial-implicit points as sparse neural key points. In contrast, topological maps provide a high-level structure that connects these local maps in compact representation.

\subsection{Navigation in Dynamic Environments}
Classical navigation uses Model Predictive Control (MPC) \cite{xiao_learning_2022} to predict future states and optimize control outputs. While MPC can be effective for some applications, it requires accurate system models, which become challenging for indoor robots operating in highly complex environments. In such environments, motion commands are often translated into incorrect real-world motion. Similarly, the Dynamic Window Approach (DWA) \cite{missura2019predictive} replies on velocity commands to estimate the direction toward the goal. This method involves local optimization with heavy reliance on an accurate motion model. While DWA's computational cost is efficient compared to MPC, it may get stuck in local minima. Our approach is to implement a novel method based on Radpid-exploring Random Tree Star (RRT*) \cite{xu2024recent} to generate adaptive global navigation plans with help from the topological map while being aware of dynamic obstacles in local frames, effectively working on partially unexplored environments.
\section{System Overview}

\begin{figure}[h]
    \centering
    \includegraphics[width=0.99\linewidth]{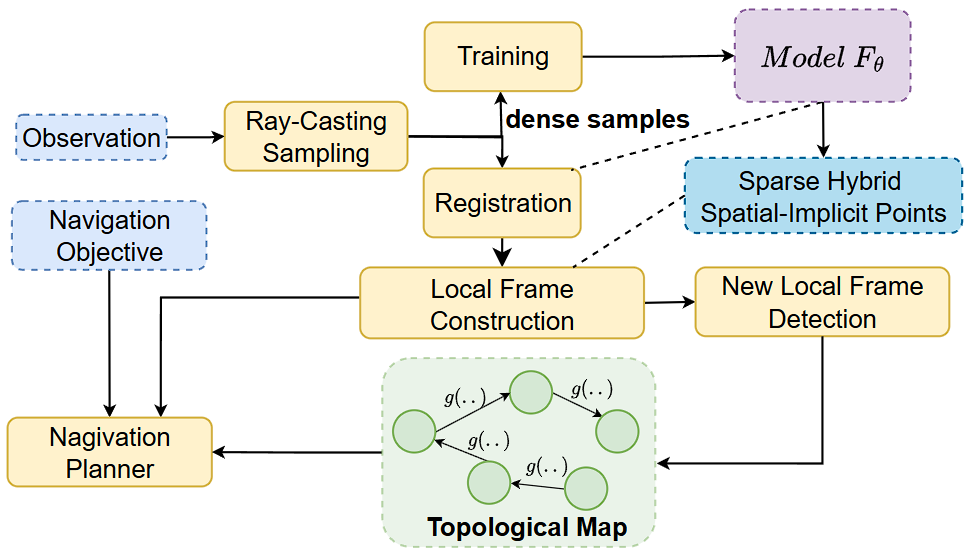}
    \caption{System Overview}
    \label{fig:data}
    \vspace{-10 pt}
\end{figure}

In the proposed system, we begin by constructing sparse local maps using map-specific implicit spatial features. To do this, we begin by sampling 3D points from a sparse representation of the surface obtained from robot sensors to create a denser and more diverse training representation. This process entails sampling points along rays originating from both the front and back surfaces and selecting points close to these surfaces. Next, we train a neural network, model \( f_\theta \), to learn the surface representation from these dense samples. After training the model, we can compute a latent vector for each point using the back-propagation process. We then re-sample these dense observations and combine them with the latent vector to create a sparse hybrid spatial-implicit representation. Following this, we construct local frames, which require performing registrations for all observations except the first one. To form the global topological structure, we differentiate one local frame from another by calculating the variance in data observed within each frame. Specifically, we assess the variance in viewpoints and travel distances to identify new local frames. Each local frame is a node on a topological map with arcs indicating the relative poses between two local frames. To develop an optimized navigation plan, we propose a novel algorithm based on RRT* that integrates both spatial-implicit local frames and topological map.
\section{Methodology}
\subsection{Problem Formulation}

The objective of constructing maps is to determine the most probable map \( m \) based on observed data \( d = \{o_1, ..,o_T\} \):

\vspace{-10 pt}
\begin{align}
\begin{split}
m^* &= \underset{m}{\textbf{argmax}} \: P(m|d) \\
    &= \underset{m}{\textbf{argmax}}\int P(m|\xi,d)P(\xi|d)d\xi
\end{split}
\label{eq:map1}
\end{align}

\noindent where \(\xi=\{\xi_1,..,\xi_T\} \) is the set of all poses. Following the derivation in \cite{thrun1998integrating}, \( m^* \) can be expressed as

\vspace{-5 pt}
\begin{equation}
\underset{m}{\textbf{argmax}} \int \prod_{t=1}^T P(o_t|m,\xi_t) \prod P(\xi_{t+1} | u_t, \xi_t) d\xi 
\label{eq:map2}
\end{equation}

In this context, translating motion commands \( u_t \) into actual motion typically is inaccurate, especially in complex environments, leading to significant errors in estimating the motion model \( P(\xi_{t+1} | u_t, \xi_t) \). Furthermore, in our approach, the motion command \( u^t \) is not particularly helpful, as we can predict the subsequent pose \( \xi_{t+1} \) based solely on the previous pose \( \xi_t \) and the current velocity vector. Consequently, we can simplify the equation by setting the motion model to 1, resulting in:

\vspace{-5 pt}
\begin{equation}
m^* = \underset{m}{\textbf{argmax}} \int \prod_{t=1}^T P(o_t|m,\xi_t) d\xi
\label{eq:map3}
\end{equation}

Equation \ref{eq:map3} shows that estimating the map is to find the most likely map such that it maximizes the perceptual model \( P(o_t|m,\xi_t)\). In this application, we utilize an RGB-D camera to capture environmental observations, including RGB and depth images, represented as \( o_t = \{I_t, D_t\} \). For each observation \( o_t \), we reconstruct a color point cloud \( P_t = \{(p_j, c_j) \in \mathbb{R}^6\} \). In this representation, \( p_j = (x_j, y_j, z_j) \in \mathbb{R}^3 \) denotes the 3D coordinates of a point, while \( c_j = (r_j, g_j, b_j) \in \mathbb{R}^3 \) indicates the color of that point relative to the current camera frame. The reconstruction uses the intrinsic camera parameters \( (focal_x, focal_y, center_x, center_y) \), as follows: \(z_j = D_t(u, v), \quad x_j = \frac{(u - center_x) z_j}{focal_x}, \quad y_j = \frac{(v - center_y) z_j}{focal_y}, \quad c_j = I_t(u, v)\). Here, \( (u, v) \) represent the pixel coordinates in the depth image \( D_t \) and the color image \( I_t \). Therefore, Equation \ref{eq:map3} becomes:

\begin{equation}
m^* = \underset{m}{\textbf{argmax}} \int \prod_{t=1}^T P(\{(p_j, c_j), j \in \mathbb{N} \}_t|m,\xi_t) d\xi 
\label{eq:map4}
\end{equation}

%===========================================================
% Next step: Think about self-learning with GNN
%===========================================================
\subsection{Learning Spatial Surface Representation Using Self-Supervise Learning}

\begin{figure}[h]
    \vspace{-10 pt}
    \centering
    \includegraphics[width=0.92\linewidth]{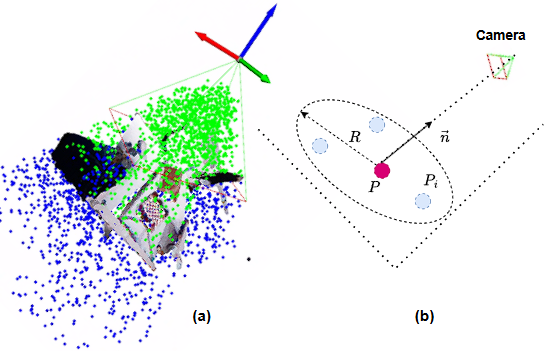}
    \caption{(a) In-distribution and out-distribution data generation. (b) Plane-radius sampling for points on the surface}.
    \label{fig:data}
    \vspace{-10 pt}
\end{figure}

We define a map as a set of 3D points with the following learning features: 
\begin{equation}
    m = \{ m_j=(p_j, f_j, s_j) \: | \: p_j \in \mathbb{R}^3, f_j \in \mathbb{R}^m, s_j \in \mathbb{N}\}
\end{equation}

Here, \(f_j\) represents the per-point learned features generated through interpolation with the features of neighboring points, assigning more weight to the points closest to the query point \(p_j\), and \( s_j\) is the semantic label. Let \(\mathcal{N}_j\) denote the neighbors of \(p_j\), and \(w_k\) be the weight for each neighbor. The feature interpolation can be expressed as:

\begin{align}
\begin{split}
    f_j &= \sum_{k } \frac{w_k}{w} \cdot f_k \\
    w_k &= \frac{1}{\|p_k - p_j\|_2} \quad \text{where } p_k \in \mathcal{N}_j, w = \sum_{k} w_k
\end{split}
\label{eq:interpolation}
\end{align}

We aim to develop a model that approximates the perceptual model \(P(o_t|m,\xi_t)\). This perceptual model predicts whether a set of color points belongs to the surface. Similar to DeepSDF \cite{park_deepsdf_2019}, we utilize a neural network denoted as \(F_\theta \) which is parameterized by \(\theta\) to predict the signed distance function (SDF) values, and  \(\phi(F_\theta) \approx 1 - P(o_t|m,\xi_t)\) where $\phi()$ is sigmoid function. We frame the learning process as a regression problem, contrasting it with approaches that view learning as a classification problem. The learned features \(f_i\) are extracted using the backpropagation method and stored outside the network. 

To train the network with the new observation \(o_t\), we begin by casting rays from the camera's perspective. For each point \(p \in P_t\), we sample points along each ray using a ratio, $l$, of the observed distance of $p$ in three ways: (1) on the surface of the point (or nearby \(p\), where the signed distance function (SDF) is approximately zero), (2) from the camera to the front of the surface, and (3) from behind the surface, as illustrated in Figure \ref{fig:data}. When sampling points in front of the surface, we select \(l\) in the range \( (0.3, 0.99) \) to ensure these points are not too close to the camera but still relatively near. For points behind the surface, we select \(l\) in the range \( (1.01, 1 + \l_b) \) where $l_b$ defines an upper limit to ensure points are not sampled beyond the object that the surface belongs to. Sampling is done using the equation: 
$ ray(l) = e + l \cdot \Vec{ep} $ where \(e\) is the camera position, $p$ is a point on the surface, and $\l_b$ is distance ratio. To sample points on the surface, we assume that the surface is perpendicular to the normal vector defined as the line from the surface point to the camera. We construct a plane using this normal vector and the surface point, then sample points on this plane within a small circular radius \(r\). We denote points on the surface as in-distribution data and points in front or behind the surface as out-distribution data. Therefore, this sampling strategy allows the network to learn from in-distribution and out-of-distribution data, facilitating self-supervised learning. 

We frame network input as a vector, including 3D coordinates, colors, and interpolated features, where 3D coordinates and colors are from sampling points while interpolated features are calculated from Equation \ref{eq:interpolation}.

\begin{equation}
    \hat{sdf} = F_\theta(p, c, f)
\end{equation}

In line with the approach described in \cite{pan_pin-slam_2024}, we employ two loss functions: (1) binary cross-entropy loss, referred to as $\mathcal{L}_{B}$, and (2) Eikonal regularization loss, denoted as $\mathcal{L}_{E}$. The binary cross-entropy loss, $\mathcal{L}_{B}$, directs how the network learns the signed distance function (SDF) values. To ensure this, we convert the SDF values to a range of $(0,1)$ using the sigmoid function $\phi_j=\phi(sdf_j)$ within $\mathcal{L}_{B}$. Meanwhile, the Eikonal loss serves to enforce smoothness on the surface. The total loss is calculated as the sum of these two losses, each weighted by their respective coefficients, $\lambda_b$ and $\lambda_e$.

\vspace{-10 pt}
\begin{align}
\begin{split}
    \mathcal{L}_{B} &= \frac{1}{N}\sum_{j=1}^{N} [ \hat{\phi}_j log\phi_j + (1-\hat{\phi}_j)log(1-\phi_j) ] \\
    \mathcal{L}_{E} &= \frac{1}{N}\sum_{j=1}^{N}(\| ( \frac{\partial F_\theta}{\partial p_{j,x}}, \frac{\partial F_\theta}{\partial p_{j,y}}, \frac{\partial F_\theta}{\partial p_{j,z}})  \|_2 - 1)^2 \\
    % \mathcal{L} &= \lambda_b\cdot\mathcal{L}_{B} + \lambda_e\cdot\mathcal{L}_{E}
\end{split}
\end{align}

We use strife sampling to generate training data supporting continual learning by incorporating new and previous observations. To prevent significant changes in model weights, a phenomenon known as catastrophic forgetting in continual learning, we implement Elastic Weight Consolidation (EWC) \cite{kirkpatrick2017overcoming} to adjust important parameters when introducing new observations.

\begin{align}
% \begin{split}
    \small
    L_{EWC} = \sum_i G_i(\theta_i - \theta_i^*)^2
    \text{, } G(\theta) &= \mathbb{E}[\frac{\partial log p(d |\theta)}{\partial \theta}]
% \end{split}
\end{align}

Summing all losses over their weights yields the final loss
\begin{equation}
    \mathcal{L} = \lambda_b\cdot\mathcal{L}_{B} + \lambda_e\cdot\mathcal{L}_{E} + \lambda_{EWC}\cdot\mathcal{L}_{EWC}
\end{equation}

We use an adaptive training loop that automatically detects when training has converged, allowing us to stop the process. This approach focuses on complex samples while accelerating the training for simpler ones. Ultimately, we can update the learned features \( f_j \) through backpropagation, applying a learning rate \( \lambda \) in a gradient descent manner.
\begin{equation}
f_j = f_j - \lambda\frac{\partial \mathcal{L}}{\partial f_j}    
\end{equation}

%===========================================================
%===========================================================
\subsection{Building Spatial-Implicit Local Frames}
\label{sec:local_frame}

\begin{figure}[h]
    \centering
    \includegraphics[width=0.99\linewidth]{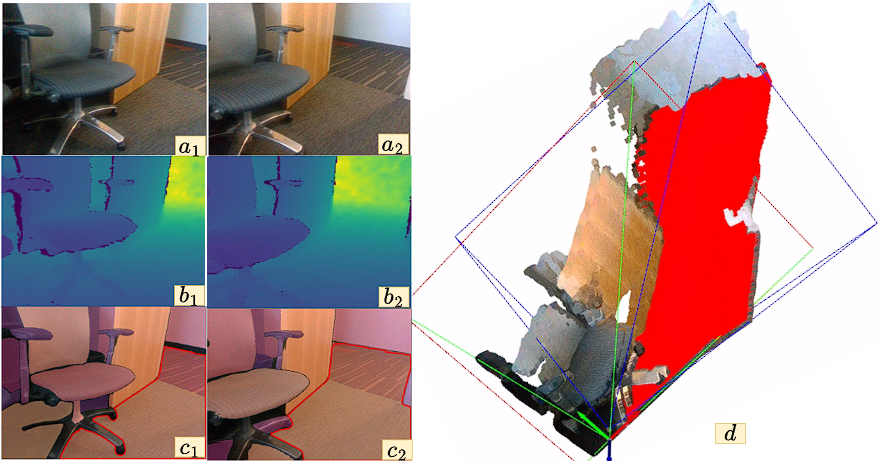}
    \caption{Building a 3D local frame from real-world office environments with semantic meaning:  ($a_1-a_2$) RGB images, ($b_1-b_2$) depth images, ($c_1-c_2$) identified traversable region on the floor (red polygons), (d) 
integration into the 3D local frame with the traversable region (red).}
    \label{fig:localframe}
    \vspace{-5 pt}
\end{figure}

The objective of this approach is to establish multiple local frames, with each local frame \(m^i\) containing several registered point clouds collected from time \(T^i_s\) to time \(T^i_e\). We aim to estimate \(m^{i*}\) based on Equation \ref{eq:map4} as:

\begin{equation}
m^{i*} = \underset{m^i}{\textbf{argmax}} \int \prod_{t=T^i_s}^{T^i_e} P(\{(p_j, c_j), j \in \mathbb{N} \}_t|m,\xi_t) d\xi 
\label{eq:localmap}
\end{equation}

For each local frame \(m^i\), we need to integrate an observation \(o_t\), which is represented as a new colored point cloud \(P^t = \{(p_j, c_j) \in \mathbb{R}^6\}\). We assume that, once properly aligned, both point clouds can be viewed as complete surfaces. Before proceeding, we address the challenge of localizing the robot using the local frame \(m^i\). Specifically, we estimate the camera pose \(\xi_t\) as modeled by $P(\xi_t | o_{T_{s:t}^i}, m^i)$. Particularly, we transform \(P^t\) to the local frame \(LF_i\) as \(P_i^t = TP^t\), where \(T \in SE(3) \subset \mathbb{R}^{4 \times 4}\).  Let \(\hat{sdf}\) represent the SDF value for point \( (p, c) \in P^t_i\) predicted by \(F_\theta\), such that \(\hat{sdf} = F_\theta(p, c, f)\), and \(sdf\) is the true surface value. Here, the \(sdf\) value equals zero since the point \(p\) lies on the surface. We define the error function as follows:

\begin{equation}
r = \hat{sdf} - sdf = \hat{sdf} = F_\theta(p, c, f) 
\end{equation}

Our goal is to minimize the error by optimizing the camera pose with local frame $LF_i$:

\begin{equation}
    T^* = \underset{T}{\text{argmin}} \sum_{p \in P_t^i} [TF_\theta(p)]^2
\end{equation}

Rather than optimizing \(T\ \in \mathbb{R}^{4 \times 4}\), we optimize the camera pose using Lie Algebra in T's tangent space, where $\xi \in \mathbb{R}^6$ and $\xi^\wedge$ is the skew-matrix of $\xi$:

\begin{equation}
    T^* = exp(\xi^{*\wedge}) =\underset{\xi}{argmin} \sum_{p \in P_t^i} [exp(\xi^\wedge)F_\theta(p)]^2
\end{equation}

% where $\xi = (p \times g, g) \in \mathfrak{se}(3) \subset \mathbb{R}^6$ and 

We define gradient vector $g$ as follows: 

\begin{equation}
g=\frac{\partial F_\theta}{\partial p}=[\frac{\partial F_\theta}{\partial p_x}, \frac{\partial F_\theta}{\partial p_y}, \frac{\partial F_\theta}{\partial p_z}]
\end{equation}

Each gradient \( g \) represents the deviation from a new point in \( P^t \) to the surface. This indicates the translation vector $tr$ from point \( p \) to the surface. The product \(a = p \times g \) forms an axis-angle representing the rotation needed to align with the target point. We aim to minimize translation and rotation regarding the observation \( P^t \) by performing Levenberg-Marquardt optimization. The Jacobian matrix is defined by:

\begin{equation}
    J = \begin{bmatrix} \frac{\partial F_\theta}{\partial a}, \frac{\partial F_\theta}{\partial tr} \end{bmatrix} = \begin{bmatrix} (p \times  g)^T, g^T \end{bmatrix}  
    % H &= JJ^T \\
    % \Delta \xi &= -(H+ \lambda I)^{-1}J^Tr
\end{equation}

The update step would be calculated by :
\begin{equation}
    \Delta \xi = -(JJ^T + \lambda_{reg} I)^{-1}J^Tr
\end{equation}

We incorporate semantics to each point in local frames to allow the robot to differentiate between the traversable regions and other objects by leveraging the 3D perception based on 2D segmentation networks \cite{tian2025yolov12, kirillov2023segment, dang2024v3d,nguyen2024volumetric} and refine noise segments using 3D spatial information.

%===========================================================
%===========================================================
\subsection{Building Topological Map}

\label{sec:topologicalmap}

To represent the global map, we build a topological map based on Equation \ref{eq:map4} as:

\vspace{-10 pt}
\begin{equation}
m^* = \{\underset{m_i}{\textbf{argmax}} \prod_{t=1}^T P(P_t|m_i,\xi_t)\}
\label{eq:topologicalmap}
\end{equation}

We utilize a perceptual model represented as \( P(P_t | m_i, \xi_t) \), where \( P^t = \{(p_j, c_j), j \in \mathbb{N}\} \). Here, \( m_i \) denotes a set of points in local frame $LF_i$, and \( \xi_t \) is identified in the previous step described in section \ref{sec:local_frame}. The sequence of observations \( P^{1:T} \) is divided into multiple segments. Each segment of observations, \( P^{T_i:T_{i+1}} \), contributes to the creation of a local frame \( LF_i \). We distinguish local frames based on variances in translation and viewpoint. An arc represents the relative pose between two local frames $i, j$, denoted as $g(LF_{i}, LF_{j}, o_{k})$, for a set of observations $o_k$. When creating a new local frame, the robot checks if it previously visited the current local frame by performing a similarity validity between spatial-implicit key points in the previous local frames and the current one. If the local frame passes this validity, it will be added as a new node into the topological map.

%===========================================================
%===========================================================
\subsection{Navigation With Local Frames And Topological Map}

In earlier sections, we discussed how to perceive the spatial surfaces of the environment using the model \(F_\theta\). Robots can use spatial-implicit points in local frames to localize themselves and infer the direction of the goal using topological maps extracted from factor graphs while using semantic information mentioned at Sec. \ref{sec:local_frame} to avoid obstacles and for other high-level tasks. Additionally, we have developed a novel \(RRT^*\)-based algorithm that integrates with a topological map, enabling navigation toward goals while maintaining both local and global awareness of the environment. Instead of sampling the entire space as traditional \(RRT^*\) algorithms do, we focus on sampling points based on the direction vector toward the goal, referred to as the goal vector. However, sampling along this goal vector can cause the robot to get stuck when faced with considerable obstacles. To address this, we generate multiple alternative direction vectors from the goal vector, ensuring that these alternative rays form a miniature angle \(\alpha\) with the goal vector and sample points along these alternative rays.
\section{Evaluation }
\begin{figure*}[t]
    \centering
    \includegraphics[width=0.99\linewidth]{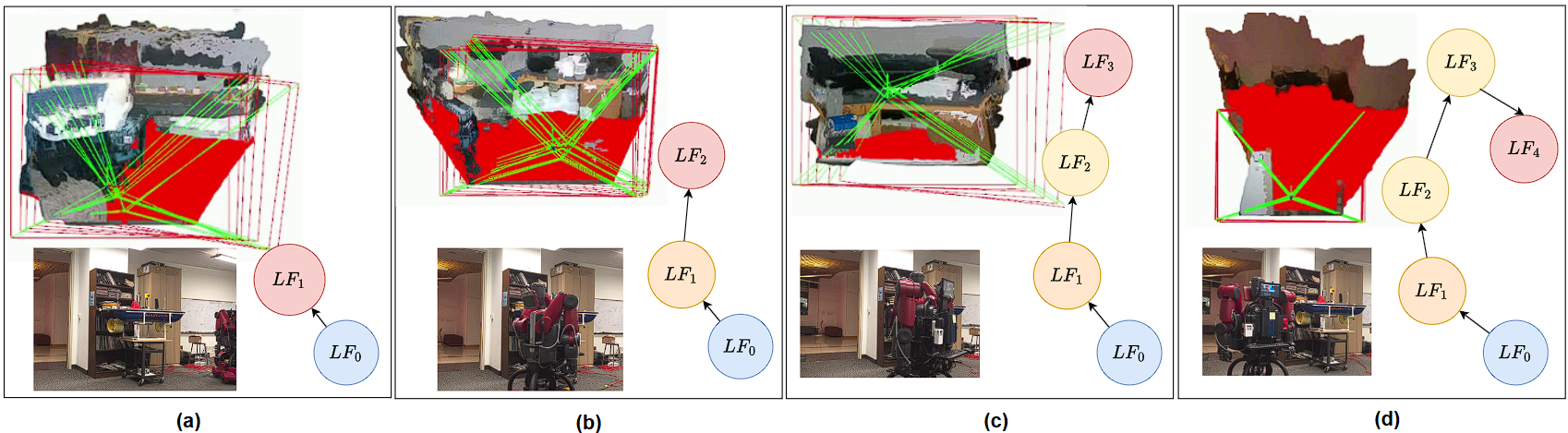}
    \caption{Qualitative results of constructing local frames and topological map from multiple observations: (a) variance in viewpoints, (b) variance in translation, (c) variance in viewpoints and translation, (d) finding the exit.}
    \label{fig:topologicalmap}
    \vspace{-14 pt}
\end{figure*}

\subsection{Evaluation Metric}
Since we focus on developing a flexible and efficient map representation for navigation while relaxing from the metric map as a conventional SLAM system to topological maps, we compare our results with iMAP \cite{sucar2021imap}, NICE-SLAM \cite{zhu_nice-slam_2022}, and ESLAM \cite{johari_eslam_2023} for local metric map. Similar to these works, we use Mean Root Square Error (MRSE) on Absolute Trajectory Estimation (ATE) as the evaluation metric. 

\subsection{Hyper-Parameter Selections}
We model our neural network using either a Multi-Layer Perceptron (MLP) or a transformer architecture to learn surface features and represent spatially implicit key points. Our experiments demonstrate that the MLP-based model significantly outperforms the transformer-based model in both accuracy and runtime. Specifically, the MLP-based models achieve twice the accuracy and are five times faster than the transformer-based models. One explanation is that the MLP-based model can capture an unordered set of points, while transformer-based models excel at sequence data. Detailed selections of hyperparameters are presented in Table. \ref{tab:parameters}.

\begin{table}[h]
    \centering
    \begin{tabular}{ c | c  c } 
    \hline
    \textbf{Notation} & \textbf{Description} & \textbf{Value} \\
    \hline
    $\lambda$ & Learning Rate & 0.001 \\
    % \hline
    $\lambda_b$ & Binary Cross Entropy Loss Weight & 1.0 \\
    % \hline
    $\lambda_e$ & Eikonal Loss Weight & 0.5 \\
     $\lambda_{EWC}$ & Elastic Weight Consolidation Weight & 0.1 \\
    % \hline
    $\lambda_{reg}$ & Damping factor Levenberg-
Marquardt & 0.001 \\
    $thres_{l}$ & Adaptive Training Loop & 0.0001 \\
    \hline
    \end{tabular}
    \caption{Hyper-parameters using in evaluation}
    \label{tab:parameters}
    % \vspace{-10 pt}
\end{table}

\subsection{TUM Dataset}

\begin{table}[h]
    % \vspace{-5 pt}
    \centering
    \begin{tabular}{ c | c  c  c } 
    \hline
     & fr1/desk & fr2/xyz & fr3/office \\
    \hline
    iMAP \cite{sucar2021imap} & 4.90 cm&  2.05 cm &  5.80 cm \\
    NICE-SLAM \cite{zhu_nice-slam_2022} & 2.85 cm &  2.39 cm &  3.02 cm \\
    ESLAM \cite{johari2023eslam} & 2.47 cm &  \textbf{1.11} cm &  2.42 cm \\
    % Our approach global map  & 38.50 &  2.39 &  3.02 \\
    Our approach  & \textbf{1.61} cm &  1.54 cm &  \textbf{1.99} cm \\
    \hline
    \end{tabular}
    \caption{Quantitative comparison of our proposed method with other methods on TUM RGB-D Dataset using ATE RSME metric}
    \label{tab:ATE}
    % \vspace{-10 pt}
\end{table}

Using the TUM dataset \cite{sturm2012benchmark}, we achieve improved or competitive precision in constructing local frames compared to the comparison methods by not concentrating on creating consistent global maps, as demonstrated in Table \ref{tab:ATE}. The algorithm utilizes an average of 120 frames to build a single local frame, generating five local frames using approximately 595 observations in the fr1/desk sequence. We apply the same method for the fr2/xyz and fr3/office sequences. To facilitate comparison, we compute the Root Mean Square Error (RMSE) for all local frames to obtain an approximate RMSE. Although there is some drift between the local frames, this will be mitigated when the robot localizes itself within a specific local frame using spatial-implicit key points.

\subsection{Navigation Planing}
\begin{figure}[h]
    % \vspace{-5 pt}
    \centering
    \includegraphics[width=0.99\linewidth]{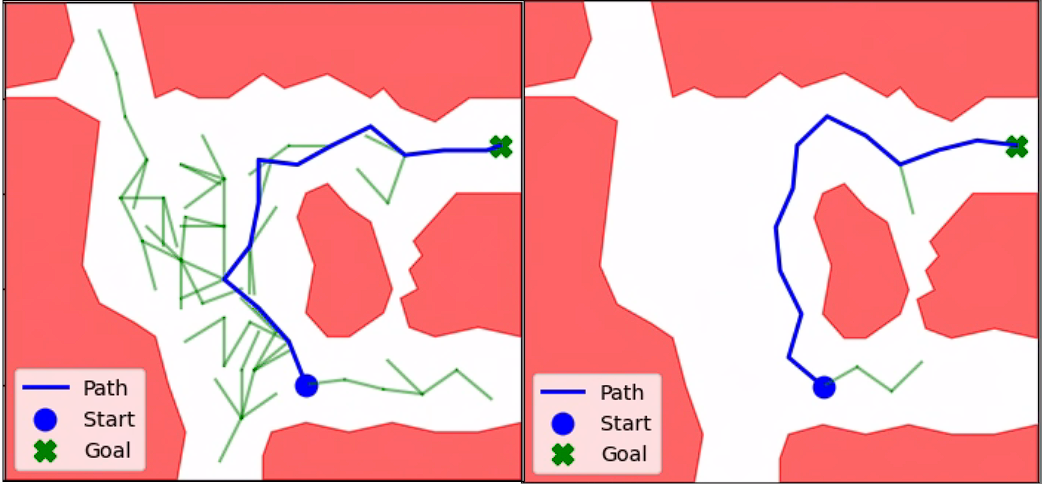}
    \caption{Left: conventional RRT* approach. Right: Our approach with local frames and topological map.}
    \label{fig:navigation}
    % \vspace{-5 pt}
\end{figure}

Whenever the robot moves, it must adjust its navigation plan to adapt to the local environment with respect to the topological map, as shown in Figure \ref{fig:topologicalmap} (in-lab experiments). We compare our navigation algorithm with the baseline method, RRT*, as shown in Figure \ref{fig:navigation}. Our approach not only generates a better route from the current location but also improves planning efficiency by reducing the exploration space. Specifically, we plan routes that are 5x faster and $50\%$ shorter than those produced by the baseline method, as shown in Table \ref{tab:nav}.

\begin{table}[h]
    % \vspace{-5 pt}
    \centering
    \begin{tabular}{ c | c  c  c }     
     \hline
      & Runtime & Travel Distance \\
      \hline
      Baseline & 110.6 ms & 6.8 m \\
      Our approach & \textbf{22.1} ms & \textbf{4.5} m \\
    \hline
    \end{tabular}
    \caption{Navigation comparison between baseline and ours}
    \label{tab:nav}
    \vspace{-10 pt}
\end{table}

\subsection{Runtime Analysis}
We benchmark different stages of map construction and navigation on a computer equipped with an Intel Core i7 processor and an Nvidia RTX 4090, 24GB VRAM graphics card. As shown in Table \ref{tab:runtime}, training a model with an input of 1,228,800 3D color points (6D points from a 640x480 RGB-D image with a dense sampling factor of 4) takes an average of 200.1 milliseconds. Theoretically, we could train the model at approximately 5 Hz. However, training the model for every single observation is not advisable, as new observations may have slight variations from the previous ones. In practice, we can train the model every five observations, allowing for a training performance of up to 25 Hz. Once the model is trained, the registration process takes an average of 40.2 milliseconds, and navigation planning is achieved in 22.1ms, illustrating that the proposed system is highly efficient and can operate in real-time.

\begin{table}[h]
    \vspace{-5 pt}
    \centering
    \begin{tabular}{ c | c  c  c }     
     \hline
     Task & Device & Runtime \\
    \hline
    Sampling (million points) & GPU & 1.2 ms \\
    Training & GPU & 200.1 ms \\
    Registration & GPU & 40.2 ms \\
    Navigation Planning & CPU & 22.1 ms \\
    
    \end{tabular}
    \caption{Runtime Analysis}
    \label{tab:runtime}
    \vspace{-15 pt}
\end{table}

\subsection{Demonstration}
The demonstration video illustrates our method's deployability on the Baxter mobile robot. We equip the robot with an Intel RealSense D435i RGB-D camera on its head. The video is available at \url{https://youtu.be/GKoSBlGPD_s}.
\section{Conclusions}
We present a bio-inspired mapping and navigation system that utilizes the latest advancements in continual learning for navigating mobile cobots through unseen 3D environments. Instead of creating an accurate but costly global map, our approach focuses on constructing local frames using hybrid spatial-implicit points for obstacle avoidance and other high-level tasks. We then integrate these local frames into a topological map structured as a factor graph for global navigation guidance. When navigating dynamic environments, the robot focuses on building these local frames to effectively avoid collisions while leveraging compact global information from the topological map to reach its destination. Additionally, we conduct experiments using real-world datasets and laboratory settings to validate our methods. Our results demonstrate that the robot can successfully navigate itself in various environments while efficiently constructing spatial-implicit local frames in real-time.

\newpage

\bibliographystyle{IEEEtran}
\bibliography{IEEEabrv, 09_references}

\end{document}